\documentclass[]{fairmeta}

\title{Towards Direct Latent-Space Synthesis for Parallel Branches in LLM-Agent Workflows}

\usepackage{xspace}
\usepackage{amsmath}
\usepackage{adjustbox}
\usepackage{tabularx}
\usepackage{multirow}
\usepackage{multicol}
\usepackage{caption}

\usepackage[most]{tcolorbox}
\usepackage{xcolor}

\usepackage[most]{tcolorbox}
\usepackage{xcolor}

\definecolor{promptblue}{RGB}{96, 165, 250}      
\definecolor{prompttitleblue}{RGB}{219, 234, 254} 
\definecolor{promptbg}{RGB}{248, 251, 255}       

\newtcolorbox{promptbox}[1]{
  enhanced,
  breakable,
  colback=promptbg,
  colframe=promptblue,
  colbacktitle=prompttitleblue,
  coltitle=black,
  fonttitle=\bfseries,
  title=#1,
  arc=2pt,
  boxrule=0.8pt,
  left=6pt,
  right=6pt,
  top=6pt,
  bottom=6pt,
  before upper={
    \setlength{\parindent}{0pt}
    \setlength{\parskip}{0.35em}
  }
}

\definecolor{casepink}{RGB}{219,39,119}
\definecolor{casepinktitle}{RGB}{252,231,243}
\definecolor{casepinkbg}{RGB}{255,247,251}
\definecolor{casesoftgray}{RGB}{248,250,252}

\newtcolorbox{casebox}[1]{
  enhanced,
  breakable,
  colback=casepinkbg,
  colframe=casepink,
  colbacktitle=casepinktitle,
  coltitle=black,
  fonttitle=\bfseries,
  title=#1,
  arc=2pt,
  boxrule=0.8pt,
  left=6pt,
  right=6pt,
  top=6pt,
  bottom=6pt,
  before upper={
    \setlength{\parindent}{0pt}
    \setlength{\parskip}{0.35em}
  }
}

\newtcolorbox{evidencebox}[1]{
  enhanced,
  breakable,
  colback=white,
  colframe=casepink!85!black,
  coltitle=black,
  fonttitle=\bfseries,
  title=#1,
  arc=2pt,
  boxrule=0.6pt,
  left=6pt,
  right=6pt,
  top=5pt,
  bottom=5pt,
  before upper={
    \setlength{\parindent}{0pt}
    \setlength{\parskip}{0.25em}
  }
}

\newcommand{\proj}{Parallel-Synthesis\xspace}
\newcommand{\textsyn}{Text-Serialization\xspace}

\author[1]{Shikun Liu}
\author[1]{Mufei Li}
\author[2]{Dongqi Fu}
\author[1]{Haoyu Wang}
\author[2]{Yinglong Xia}
\author[2]{Hong Li}
\author[2]{Hong Yan}
\author[1]{Pan Li}

\affiliation[1]{Georgia Institute of Technology}
\affiliation[2]{Meta}

\abstract{Large language models increasingly serve as execution engines for agentic systems, yet they still consume context through a sequential text interface. This creates a mismatch with modern structured agent workflows, in which independent branches explore subtasks, retrieve evidence, or generate candidate solutions before a final synthesis step. Existing systems typically merge these branches by concatenating their textual outputs, which discards the parallel structure and incurs redundant prefill computation. In this work, we introduce \proj, a plug-and-play framework that enables a synthesizer to directly consume the KV caches produced by parallel worker agents. \proj combines a cache mapper that calibrates independently generated branch caches with a fine-tuned synthesizer adapter that enables generation from this non-sequential cache interface. We train \proj using data that exposes the synthesizer to parallel cache contexts, teaches aggregation across cached branches, and distills reasoning behavior from standard text-concatenation-based synthesis. Across nine downstream datasets spanning math, science QA, code generation, GAIA, and multi-agent database diagnosis, \proj matches or outperforms text-based synthesis on seven datasets and remains close on the other two. It also reduces time-to-first-token by 2.5×--11×, suggesting that direct cache-based synthesis is a promising interface for more native and efficient synthesis over parallel agent branches.}

\date{\today}
\correspondence{\email{shikun.liu@gatech.edu}, \email{panli@gatech.edu}}

\begin{document}

\maketitle

\vspace{-1mm}
\section{Introduction}
\label{sec:intro}
Large language models (LLMs) serve as the core execution engines for agentic systems due to their advanced reasoning and tool-use capabilities~\citep{yao2022react, wang2024survey, schick2023toolformer}. Despite these advances, LLMs still operate through a sequential context interface: the entire agent state, including message history, tool outputs, retrieved evidence, and intermediate reasoning, is serialized into a single ordered token prefix for next-step execution. While this interface has been highly effective for scaled-up autoregressive training, it provides only an indirect way to represent the rich dependency structure that arises in agentic computation. Moreover, as histories grow, such serialization also imposes artificial dependencies among otherwise unrelated information, lengthening the prefix that needs to be processed and making it harder for the model to focus on the evidence most relevant to the current query~\citep{liu2024lost, shi2023large, zhang2024chain}.

This mismatch becomes especially clear when agent workflows contain parallel structure. Many complex tasks are better represented as directed acyclic graphs (DAGs), where independent subtasks, retrieval branches, candidate solutions, or tool calls can be executed separately and merged later. Such parallel-then-synthesize patterns have emerged as an effective orchestration strategy for improving both reliability and efficiency in complex problem solving. 
Even in single-agent settings, mathematical reasoning and code generation systems often improve reliability by sampling multiple test-time trajectories and then selecting among candidate solutions or aggregating sampled reasoning chains~\citep{chen2021evaluating,brown2024large,wangself}.
In multi-agent systems, this structure becomes more explicit: parallel sub-agents can gather evidence from different perspectives or solve independent subtasks before a downstream agent merges their findings~\citep{anthropic2025multiagentresearch, chai2025scimaster, li2025parallelmuse, openai2026codexsubagents, anthropic2026claudecode_subagents}.

However, because LLMs natively consume only linearized contexts, synthesizing multiple independent branches typically requires concatenating their outputs into a single expanded text prompt for the synthesizer. This text-based communication interface discards the original DAG structure of the computation and incurs redundant prefill cost by re-encoding information that has already been processed during worker decoding. Since such non-sequential agent workflows become increasingly common, it is timely to move beyond treating context as a single sequential prefix and to consider interfaces that preserve independence among branches while supporting efficient and accurate synthesis.

One natural engineering mitigation is to summarize each branch and pass the summaries to the synthesizer~\citep{anthropic2026claudecode_subagents, sun2025scaling, kang2025acon}. While this can reduce the long-context burden, it still relies on natural-language communication: summaries must be generated, serialized, and re-encoded through prefill, and the summarization step may discard details that are important for downstream judgment. In this work, we study a more direct alternative, one that points toward a more fundamental agent capability: can a synthesizer agent directly consume the latent states of parallel worker branches, rather than relying on concatenated plain-text outputs? Such an interface could reduce redundant computation while better preserving the parallel structure of non-sequential agent workflows.

Communicating through KV caches instead of explicit text is related to recent work on latent-space communication between agents. Most existing formulations focus on one-to-one latent transfer, where a latent state or cache produced under one context is passed to another agent~\citep{zou2025latent, yekvcomm, fu2025cache}. Our setting instead studies many-to-one synthesis over parallel agent branches, where the downstream synthesizer must jointly consume and integrate several independently produced branch states. Another adjacent line of work studies parallel encoding and cache reuse for long-context RAG, where retrieved document chunks are independently encoded into KV caches and later reused during generation~\citep{yangkvlink, yangape, ratner2023parallel, yao2025cacheblend}. While related in mechanism, this setting differs from ours in how cached contexts are constructed, how KV states are obtained, and what the downstream model is required to do. We defer a more detailed discussion to Sec.~\ref{sec:parallel_encoding}.


In this work, we present \textbf{\proj}, a plug-and-play synthesizer framework for direct synthesis over the cached states of parallel agent branches. Unlike standard text-based synthesis, which concatenates branch outputs and prefills their raw text again, \proj enables the downstream synthesizer to reuse caches produced by parallel worker agents for subsequent generation. Since this departs from the standard serialization assumed by standard LLM pipelines, \proj employs a cache mapper to calibrate independently produced branch caches, together with a synthesizer LoRA that adapts the model to this non-sequential cache interface. To support this adaptation, we construct training datasets with complementary goals: 1) large-scale continued-pretraining-style dialogue data for broad adaptation to parallel cache contexts; 2) multi-source synthesis tasks that teach the model to aggregate information from multiple cached contexts when responding to queries; and 3) distillation data from a standard text-concatenation-based synthesis pipeline to strengthen reasoning and judgment over complex agent trajectories.

We evaluate \proj on downstream tasks spanning standard problem-solving and agentic branch-and-synthesize settings, including mathematical reasoning, scientific QA, code generation, tool-use agentic QA, and multi-agent diagnostic reasoning. Across these diverse settings, \proj largely preserves or improves the quality of standard text-based synthesis while substantially improving efficiency: it matches or outperforms text-based synthesis on 7 of 9 datasets, with only small gaps on the remaining two, and reduces time-to-first-token (TTFT) by 2.5$\times$--11$\times$ by avoiding the re-prefill of branched worker outputs. Notably, \proj achieves gains on several reasoning-heavy tasks, suggesting that direct latent-state synthesis can be not only faster but also competitive, and sometimes beneficial, for integrating parallel reasoning trajectories. 


\begin{figure}[t]
\vspace{-9mm}
    \centering
\includegraphics[
        width=0.95\linewidth,
        trim={0.2cm 9cm 0.3cm 1cm},
        clip
    ]{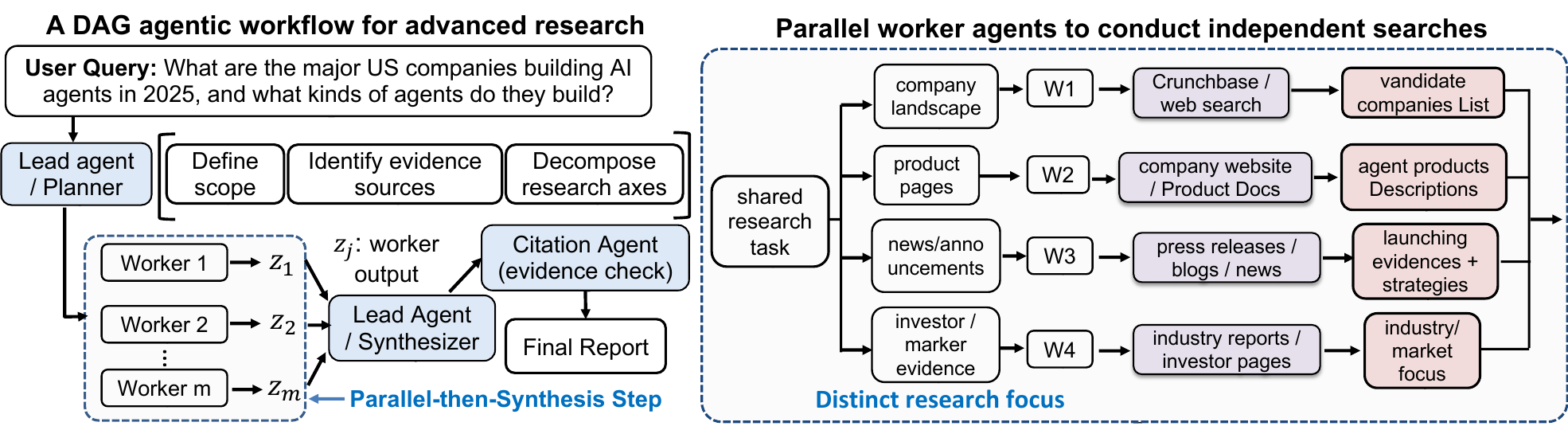}
\vspace{-1mm}
    \caption{An example of DAG agentic workflow for advanced research containing using parallel worker agents for research in different directions.}
    \label{fig:DAG}
    \vspace{-7mm}
\end{figure}
\vspace{-5mm}

\vspace{2mm}
\section{Preliminaries and Problem Formulation}

As discussed in Sec.~\ref{sec:intro} and illustrated in Fig.~\ref{fig:DAG}, modern agentic workflows often contain parallel branches executed by independent sub-agents, whose intermediate results must later be merged by a downstream synthesizer. The branching stage is naturally supported by autoregressive decoding: multiple sub-agents can decode from a shared contiguous prefix, forming a tree-structured context. The challenge arises at synthesis, where the downstream agent must condition jointly on several branches rather than continue from a single linear prefix. This merge operation turns the context topology from a tree into a DAG, exposing a mismatch with the standard single-prefix interface of LLMs. This paper studies the resulting problem: how to enable a downstream agent to synthesize information from multiple parallel branches while preserving the DAG-structured context.

\subsection{Problem Formulation: Parallel Agent Branching Synthesis}
\label{sec:problem}
We consider a general parallel-agent workflow with $m$ worker agents followed by a synthesizer. All agents are instantiated from the same autoregressive backbone model $f_\theta$. Each worker agent $j \in \{1,\ldots,m\}$ is conditioned on a branch context $c_j$ and generates an output segment $z_j \sim P_\theta(\cdot \mid c_j)$. We decompose the worker context as $c_j = c^{\mathrm{sh}} \circ c_j^{\mathrm{br}}$, where $c^{\mathrm{sh}}$ denotes the context shared across workers, $c_j^{\mathrm{br}}$ denotes the branch-specific context, and either may be empty depending on the workflow. After all workers complete, a synthesizer receives its own prompt or instruction $u$ and generates a final output $\mathbf{y}$ by conditioning jointly on the information produced by the parallel branches.

The central question is how the branch outputs $\{z_j\}_{j=1}^m$ should be communicated to the synthesizer. We compare two synthesis interfaces: the standard text-serialization interface and a cache-based interface that directly reuses the KV states produced by worker agents.

\noindent\textbf{Text serialization.}
The standard route communicates worker outputs as explicit text. Under a fixed worker ordering, the synthesizer forms a single sequential input prefix $x^{\mathrm{text}} = u \circ z_1 \circ \cdots \circ z_m$ and generates autoregressively as $y_t \sim P_{\mathrm{text}}(\cdot \mid x^{\mathrm{text}} \circ y_{<t})$. This interface is simple and directly compatible with existing LLMs. However, it flattens branch-separated outputs into a single sequence and requires the synthesizer to prefill the worker outputs again, even though these outputs have already been generated by the workers.

\noindent\textbf{Cache-based synthesis.}
Instead of communicating worker outputs as text, we study whether the synthesizer can directly consume the KV states produced during worker decoding. For worker $j$, let $\mathrm{KV}_\theta(z_j \mid c_j)=\{(K_{z,j}^{\ell}, V_{z,j}^{\ell})\}_{\ell=1}^L$ denote the layer-wise KV states associated with the output segment $z_j$ generated under context $c_j$. The synthesizer only prefills its own prompt, $\mathrm{KV}_\theta(u)=\{(K_u^\ell, V_u^\ell)\}_{\ell=1}^L$, and reuses the worker KV states without regenerating them. It then decodes using the directly-concatenated caches $K_{\mathrm{syn}}^\ell=[K_{z,1}^\ell;\cdots;K_{z,m}^\ell;K_u^\ell]$ and $V_{\mathrm{syn}}^\ell=[V_{z,1}^\ell;\cdots;V_{z,m}^\ell;V_u^\ell]$, generating tokens as $y_t \sim P_{\mathrm{kv}}(\cdot \mid \{(K_{\mathrm{syn}}^\ell,V_{\mathrm{syn}}^\ell)\}_{\ell=1}^L, y_{<t})$.

\noindent\textbf{Target.} The text-based and cache-based routes expose the same branch outputs $\{z_j\}_{j=1}^m$ to the synthesizer, but through different representations. Our goal is to make cache-based synthesis behave like the standard text-serialization route while avoiding text concatenation and redundant prefill. In particular, for likely generated output sequences $\mathbf{y}$ under the text route, we aim for
\vspace{-1mm}
\[
P_{\mathrm{kv}}\!\left(
\mathbf{y} \mid u,\{\mathrm{KV}_\theta(z_j \mid c_j)\}_{j=1}^m
\right)
\approx
P_{\mathrm{text}}\!\left(
\mathbf{y} \mid u,z_1,\ldots,z_m
\right).
\vspace{-3mm}
\]
\vspace{-4mm}


\noindent\textbf{Why this is nontrivial.} The cache-based route above goes beyond standard KV-cache reuse. Unlike ordinary autoregressive decoding, where cached states arise from continuing a single prefix, the synthesizer receives caches extracted from disjoint parallel workers, each generated under its own local context. These caches do not fit the ordinary continuation-based KV cache reuse setting for reuse at synthesis time fundamentally nontrivial.

\noindent\textbf{Covered workflow scenarios.} 
The parallel synthesis problem arises across several common agentic workflows. The main difference among these settings lies in how each worker context $c_j = c^{\mathrm{sh}} \circ c_j^{\mathrm{br}}$ is formed before the synthesizer aggregates the branch outputs.

\textit{1). Single-turn parallel problem solving.}
All workers receive the same one-shot task input, such as a math or code-generation problem, and independently generate candidate solutions. In this case, $c_j = c^{\mathrm{sh}}$ for all $j$, and workers differ only through their independently sampled outputs $z_j$.

\textit{2). Multi-turn parallel trajectory rollout.}
Workers share the same initial prompt or user query, but each worker accumulates its own reasoning, tool calls, observations, or intermediate messages over time. Here, $c^{\mathrm{sh}}$ contains the shared task context, while $c_j^{\mathrm{br}}$ contains the worker-specific trajectory. The output $z_j$ represents the final response or summary produced from that trajectory.

\textit{3). Distinct sub-task execution.}
The original task is decomposed into different sub-tasks, each assigned to a separate worker. In this case, $c^{\mathrm{sh}}$ may contain the global task description, while $c_j^{\mathrm{br}}$ contains the sub-task instruction and any branch-specific trajectory. The output $z_j$ represents the result of the corresponding sub-task branch.

\subsection{Difference from RAG-Style Parallel Encoding}
\label{sec:parallel_encoding}
Several prior works study KV-cache reuse for long-context RAG, where retrieved documents are independently encoded into cache blocks and later reused for query-conditioned generation~\citep{ratner2023parallel}. To align with our notation, let each retrieved document chunk be denoted by $z_j$. These methods encode each $z_j$ into a cache block and later reuse the collection of cached chunks during downstream decoding. Since the cached chunks are not produced as a single contiguous prefix, such methods often introduce auxiliary mechanisms, such as partial recomputation~\citep{yao2025cacheblend}, attention calibration~\citep{yangape}, or linking tokens~\citep{yangkvlink}, to make independently encoded caches jointly usable.

Our setting is related but differs in several important ways. First, in RAG-style cache reuse, $z_j$ typically denotes an isolated document chunk encoded without a worker-specific prefix context $c_j$ or prior workflow state. In parallel agent synthesis, by contrast, $z_j$ denotes the output of a worker branch generated under its own context $c_j$, which may include a local instruction, reasoning trajectory, tool interaction, or intermediate observation. Second, RAG chunks $z_j$ are often obtained by segmenting a longer document~\citep{ratner2023parallel, mablock}, and therefore may not be semantically self-contained. This can leave unresolved dependencies across chunks, requiring mechanisms that repair or calibrate attention across cached blocks. In our setting, $z_j$ is usually a more coherent unit of agent progress, such as a candidate solution, branch summary, or subtask result. This makes branch caches more naturally suited to parallel processing and downstream synthesis than arbitrary document chunks.

The downstream objective is also different. RAG-style cache reuse mainly aims to help the model retrieve and use evidence from cached chunks $\{z_j\}_{j=1}^m$ for a final query. Parallel agent synthesis instead requires the synthesizer to compare, judge, reason over, aggregate, and summarize heterogeneous worker outputs $\{z_j\}_{j=1}^m$ under a broader range of instructions. Thus, while RAG-style cache reuse provides useful technical inspiration, it targets a different communication pattern and downstream objective from the parallel agent branching synthesis setting studied in this work. Consistent with this distinction, our experiments in Sec.~\ref{sec:exp} show that directly adapting RAG-style cache-reuse methods to parallel agent synthesis leads to weak downstream performance.

\section{Methodology}

\label{sec:method}

\subsection{Model Architecture for the Parallel Synthesizer}
\label{sec:model}

\begin{figure}[t]
\vspace{-9mm}
    \centering
    \includegraphics[
        width=0.85\linewidth,
        trim={0.7cm 0.42cm 1.55cm 0.48cm},
        clip
    ]{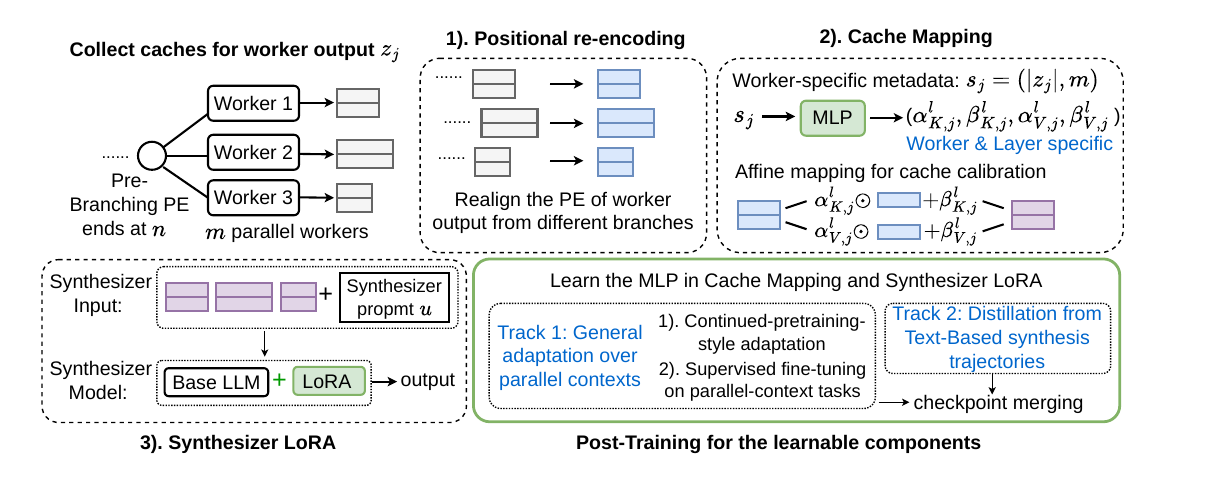}
    \vspace{-2mm}
    \caption{The model pipeline after obtaining caches from parallel worker agents together with the post-training illustration for learnable cache mapping and synthesizer LoRA.}
    \label{fig:pipeline}
    \vspace{-6mm}
\end{figure}

Our framework equips the synthesizer with three components: positional re-encoding to realign worker outputs in different branches, a cache mapper to further calibrate these reused caches before synthesis, and a synthesizer-side LoRA adaptation to help the model interpret and reason over multiple branches jointly. The overall pipeline is illustrated in Fig.~\ref{fig:pipeline}.


\noindent\textbf{Positional re-encoding.}
Each worker output $z_j$ is originally generated after the full worker-side prefix context $c_j$. Therefore, if we index the tokens in $z_j$ by their local offset $r=0,\ldots,|z_j|-1$, then the $r$-th token of $z_j$ is cached at RoPE position $|c_j|+r$. 
Suppose the agent workflow shares a common pre-branch context before spawning parallel worker agents. Let $n$ denote the first RoPE position immediately following this shared prefix at the branching point, the exact step where the original workflow paused. During synthesis, we anchor the first token of every parallel worker's output to position $n$, ensuring each trajectory structurally branches from the same continuation point rather than being serialized. Concretely, for the $r$-th token of branch output $z_j$, we reassign it to the shared target position $n+r$ as:
\vspace{-2mm}
\[
\tilde{k}_{z,j,r}^{\ell}
=
R(n+r)\,R(|c_j|+r)^{-1}k_{z,j,r}^{\ell},
\quad
\tilde{v}_{z,j,r}^{\ell}=v_{z,j,r}^{\ell}, \quad \text{for} \; r=0,\ldots,|z_j|-1
\vspace{-3mm}
\]
\vspace{-3mm}

where $R(\cdot)$ denotes the RoPE rotation operator~\citep{su2024roformer} and obtain re-encoded branch cache $(\tilde K_{z,j}^{\ell}, \tilde V_{z,j}^{\ell})$.

This positional re-encoding preserves the original token order within each worker agent's output while aligning all worker outputs to the same post-branch positional range. It therefore presents the worker outputs to the synthesizer as parallel continuations from a shared branching point, rather than as an artificially serialized sequence, which may help reduce order-induced positional bias~\citep{wanggraph, wangeliminating}.


Let $M_{\mathrm{par}}=\max_{j\in[m]} |z_j|$ be the maximum worker output length. After PE re-encoding, all worker outputs lie in the shared interval $[n,\,n+M_{\mathrm{par}}-1]$, and the synthesizer prompt $u$ is placed after this block. Decoding resumes from position $n+M_{\mathrm{par}}+|u|$, attending to $(K_{\mathrm{syn}}^\ell, V_{\mathrm{syn}}^\ell)$ at every layer.

\noindent\textbf{Cache mapping.}
After positional re-encoding, we further calibrate the reused worker caches with a learnable cache mapper before they are consumed by the synthesizer. This is related in spirit to APE~\citep{yangape}, which uses inference-time attention calibration to make independently encoded contexts better approximate sequential encoding. In contrast, we learn an explicit mapping over cached keys and values jointly with the synthesizer-side LoRA.

Concretely, we form worker-specific metadata
$
s_j = (|z_j|, m),
$
where $|z_j|$ is each worker output length and $m$ is the total number of parallel workers. We learn an MLP that takes the worker statistics $s_j$ as input and predicts, for each layer $\ell$, the affine coefficients $\alpha_{K,j}^\ell, \beta_{K,j}^\ell, \alpha_{V,j}^\ell,$ and $\beta_{V,j}^\ell$. Thus, the resulting transformations remain worker-specific through $s_j$ and layer-specific through $\ell$. We then apply them elementwise to the position-re-encoded keys and values before synthesis, which adds negligible computational overhead compared with re-prefilling worker outputs as text:
\vspace{-2mm}
\[
\hat K_{z,j}^\ell
=
\alpha_{K,j}^\ell(s_j)\odot \tilde K_{z,j}^\ell
+
\beta_{K,j}^\ell(s_j),
\qquad
\hat V_{z,j}^\ell
=
\alpha_{V,j}^\ell(s_j)\odot \tilde V_{z,j}^\ell
+
\beta_{V,j}^\ell(s_j).
\vspace{-4mm}
\]
\vspace{-4mm}

Ultimately, the synthesizer forms a unified past cache by concatenating the calibrated worker caches with the prefilled cache of its prompt:
$
K_{\mathrm{syn}}^\ell = [\hat K_{z,1}^\ell;\cdots;\hat K_{z,m}^\ell; K_u^\ell],
\,
V_{\mathrm{syn}}^\ell = [\hat V_{z,1}^\ell;\cdots;\hat V_{z,m}^\ell; V_u^\ell].
$



\noindent\textbf{Synthesizer LoRA adaptation.}
Cache calibration alone is insufficient. 
We train the synthesizer via LoRA jointly with the cache mapper. In practice, this forms a plug-and-play synthesizer-side adapter: it is activated only when parallel synthesis is used, while worker-side execution remains unchanged.

\vspace{-1mm}
\subsection{Model Training}
\label{sec:training}

A parallel cache-based synthesizer must acquire two complementary abilities. First, it should learn to \emph{read} and \emph{interpret} parallel worker caches. Second, it should preserve its \emph{reasoning} ability when using those cached trajectories under the synthesizer instruction. We therefore train two complementary post-training tracks from the same base model and merge them afterward: Track~1 provides broad and scaled-up adaptation to parallel cache contexts, while Track~2 targets realistic synthesis and reasoning over parallel agent trajectories through distillation from the text-serialization route.


\noindent\textbf{Training data format and objective.}
We denote the trainable synthesizer-side parameters by $\phi$, including the cache mapper and synthesizer LoRA, while keeping the backbone parameters $\theta$ frozen. Following the notation in Sec.~\ref{sec:problem}, each training instance is constructed as
$
(u,\{(c_j,z_j)\}_{j=1}^m,y^\star),
$
where $u$ is the synthesizer prompt, $\{z_j\}_{j=1}^m$ are the parallel worker outputs generated under their corresponding context prefixes $\{c_j\}_{j=1}^m$, and $y^\star$ is the supervision target. All training tracks optimize the same teacher-forced next-token objective under the cache-based synthesis route, differing only in how the training instances are constructed:
$
\mathcal{L}_{\mathrm{SFT}}
=
-\sum_{t=1}^{|y^\star|}
\log
P_{\theta,\phi}\!\left(
y_t^\star
\mid
u,\{\widehat{\mathrm{KV}}_{\theta,\phi}(z_j \mid c_j)\}_{j=1}^m, y_{<t}^\star
\right).
$

\noindent\textbf{Track 1: General adaptation over parallel contexts.} This training track equips the synthesizer to process independently encoded parallel contexts rather than a traditional serialized sequence. We use two complementary data types: continued-pretraining-style data and supervised fine-tuning data on explicit parallel-context tasks. The former teaches the model naturally resume generation from multi-turn histories structured as parallel KV cache. The latter provides explicit, task-oriented supervision, compelling the model to actively synthesize information across these parallel branches to resolve queries. The dataset statistics are included in Appendix~\ref{app:post-train-data}.

\textit{1) Continued-pretraining-style adaptation.} We use large-scale multi-turn dialogue data, WildChat~\citep{zhao2024wildchat}, UltraChat~\citep{ding2023enhancing}, and LMSYS-Chat~\citep{zheng2023lmsyschat1m}, to adapt the model to generate from non-sequential prefix caches. We treat the final user query as $u$ and the final assistant response as $y^\star$. The preceding dialogue turns are encoded independently as parallel caches rather than concatenated into one serialized history. Since the target response may depend only weakly on earlier turns, this stage primarily acts as an architectural warm-up that adapts the model to conditioning on parallel contexts.


\textit{2) Supervised fine-tuning on parallel-context tasks.}
We construct training samples from tasks whose inputs can be naturally decomposed into independent, self-contained semantic blocks, and whose solutions explicitly require the model to follow the instruction and read and understand the provided parallel contexts. For agentic tool-use data such as Toucan~\citep{xu2025toucan} and DTA-Tool~\citep{zhu2025dividethenaggregateefficienttoollearning}, we extract parallel tool-use trajectories. Here, $u$ is the user query, each tool-call--observation pair is treated as a parallel branch output $z_j$, the preceding agent history defines its prefix context $c_j$, and $y^\star$ is the next agent response that interprets and uses these tool results. For in-context learning data like FLAN~\citep{weifinetuned}, $u$ is the target question, $y^\star$ is the answer, and preceding few-shot exemplars serve as the parallel contexts. Finally, for multi-document QA such as 2Wiki~\citep{xanh2020_2wikimultihop}, $u$ is the query and individual source documents form the parallel branches. We construct $y^\star$ as a reasoning trace that first summarizes the provided contexts, analyzes their relevance to $u$, and then deduces the target answer.

\noindent\textbf{Track 2: Distillation from Text-Based Synthesis Trajectories.}
This track targets the richer synthesis behavior required in realistic agentic settings, where the synthesizer must reason over, compare, and judge prior sub-agent progress under detailed task instructions. Although Track~1 teaches the model to read and use parallel caches, it does not provide enough supervision for higher-level reasoning over the evidence encoded in them, which limits end-task synthesis performance in practice. We therefore distill synthesis trajectories from the text-serialization route, using them as teacher signals for reasoning-intensive synthesis.


Concretely, we use BrowseComp~\citep{wei2025browsecomp} to construct parallel worker trajectories and synthesis supervision. For each sample, we first generate three parallel worker rollouts and then run the synthesizer under the text-based synthesis route, where the worker outputs are serialized as text context. Although these synthesis traces are constructed from three worker trajectories, the trained synthesizer is not tied to this rollout count and generalizes to different numbers of parallel trajectories at evaluation time. We then collect the resulting synthesis trajectories and filter out failed or low-quality cases, retaining only those with reliable reasoning and valid final responses as supervision targets $y^\star$ for cache-based training. Each retained example is converted into our cache-based training format: for worker $j$, $c_j$ is the worker-local history before its final output turn, including prior messages, tool calls, tool observations, and intermediate reasoning, and $z_j$ is the final output communicated to the synthesizer. The synthesizer prompt $u$ is kept unchanged from the text-based synthesis route.

\noindent\textbf{Checkpoint merging.}
We train the two tracks independently and merge them afterward, since sequential tuning gives weaker results (Sec.~\ref{sec:ablation}), suggesting partial overwriting of earlier capabilities. This matches findings that continued post-training can induce forgetting, whereas merging separately specialized checkpoints better preserve complementary updates~\citep{yangmix,xiao2024lm,harmon2025mapping,hemergebench}. We merge the parameters of the cache mapper and the LoRA by weighted averaging, with $\lambda=0.5$ by default.

\vspace{-2mm}
\section{Experiments}
\vspace{-1mm}
\label{sec:exp}
\noindent\textbf{Evaluation datasets}
We evaluate \proj on nine datasets across four domains: (1) Mathematical reasoning: AIME 2024/2025~\citep{maxwell_jia_aime_2024, aime25} and GSM8K~\citep{cobbe2021gsm8k}; (2) Scientific QA: GPQA~\citep{rein2024gpqa} and MedQA~\citep{yang2025llm}; (3) Code generation: HumanEval-Plus and MBPP-Plus~\citep{liu2023your}; and (4) Multi-turn agentic benchmarks with tool use: GAIA~\citep{mialon2023gaia} and the database task from MARBLE~\citep{zhu2025multiagentbench}. In all tasks except MARBLE Database, parallel worker agents are used to explore multiple candidate solution or search trajectories. In MARBLE Database, worker agents instead handle decomposed subtasks to independently analyze possible root causes of database errors. We use 5 parallel worker agents for MARBLE Database task since it is determined by fixed sub-tasks and 3 parallel workers for the rest datasets. These benchmarks cover diverse settings where parallel-then-synthesize execution is useful. Importantly, none of them overlap with the post-training datasets, so all results reflect out-of-domain generalization. More details are provided in Appendix~\ref{app:eval-data}. 

\noindent\textbf{Models and Baselines.}
We use Qwen3-14B as the backbone for \proj and, whenever possible, for all baselines. We organize baselines into three groups. \textit{Single-trajectory and answer-level aggregation} baselines test whether gains come simply from sampling multiple workers: \textit{Single} uses one agent trajectory and its final answer, while \textit{Voting} aggregates only the final answers $\hat{y}_j$ extracted from worker outputs $z_j$ via
$
\hat{y}=\arg\max_a \sum_{j=1}^{m}\mathbf{1}[\hat{y}_j=a],
$
counting ties as incorrect. We omit these two baselines for MARBLE Database, which is inherently a decomposed multi-agent benchmark. \textit{Text-based synthesis} is represented by \textit{Text-Serialization} from Sec.~\ref{sec:problem}, which uses the same worker outputs as \proj but communicates them by concatenating their textual outputs into the synthesizer prompt. \textit{Cache-based synthesis} baselines adapt RAG-style parallel-encoding methods, including APE~\citep{yangape}, CacheBlend~\citep{yao2025cacheblend}, and KVLINK~\citep{yangkvlink}: APE calibrates attention at inference time, CacheBlend selectively recomputes KV states, and KVLINK uses linking tokens between cache blocks with a post-trained synthesizer. For KVLINK, we use the largest released checkpoint; worker trajectories are generated by Qwen3-14B and then encoded into KVLINK's linked-cache format for final synthesis. Additional evaluation details are provided in Appendix~\ref{app:training_config}.


\vspace{-1mm}
\subsection{Main Results}


\begin{table}[t]
\vspace{-9mm}
\centering
\caption{Performance across math, code generation, and scientific QA benchmarks in Accuracy. We \textbf{bold} where \proj matches or outperforms \textsyn}
\label{tab:main_problem}
\resizebox{0.85\linewidth}{!}{%
\begin{tabular}{l|ccc|cc|cc}
\toprule
\multirow{2}{*}{Method}
& \multicolumn{3}{c|}{Math problem}
& \multicolumn{2}{c|}{Code generation}
& \multicolumn{2}{c}{Scientific QA} \\

& AIME 2024 & AIME 2025 & GSM8K
& HumanEvalPlus & MBPPPlus
& GPQA & MedQA \\
\midrule
Single              & 23.33 & 13.33 & 92.42 & 84.15 & 76.19 & 31.31 & 82.72 \\
Voting              & 30.00 & 20.00 & 92.20 & 89.63 & 79.37 & 48.99 & 83.74 \\
CacheBlend                 & 13.33  & 23.33 & 79.15 & 51.22 & 52.91 & 12.63 & 40.38 \\
APE                 & 6.67  & 10.00 & 49.66 & 36.59 & 52.91 & 10.10 & 17.05 \\
KVLINK                 & 50.00  & 40.00 & 92.72 & 87.80 & 78.31 & 35.35 & 67.05 \\
Text-Serialization             & 63.33 & 23.33 & 92.80 & 90.24 & 81.75 & 50.00 & 82.72 \\
\midrule
\proj & \textbf{63.33} & \textbf{46.67} & \textbf{94.69} & \textbf{90.85} & 80.42 & \textbf{52.02} & \textbf{83.58} \\
\bottomrule
\end{tabular}%
}
\vspace{-3mm}
\end{table}
\begin{figure}[t]
\centering

\begin{minipage}[t]{0.44\linewidth}
\centering
\captionof{table}{Performance on GAIA and MARBLE. Each entry reports the number of correct samples. The total numbers of samples are 53, 86, and 26 for GAIA Levels 1--3, respectively, and 100 for MARBLE Database.}
\label{tab:main_agentic}
\resizebox{\linewidth}{!}{%
\begin{tabular}{l|ccc|c}
\toprule
\multirow{2}{*}{Method}
& \multicolumn{3}{c|}{GAIA}
& \multicolumn{1}{c}{MARBLE} \\

& Level 1 & Level 2 & Level 3 & Database \\
\midrule
Single             & 14 & 18 & 1 & N/A  \\
Voting             & 19 & 16 & 1 & N/A \\
CacheBlend                & 17 & 16 & 0 &  7 \\
APE                & 20 & 13 & 2 & 18  \\
KVLINK                & 20 & 20 & 1 & 10  \\
Text-Serialization     & 24 & 22 & 2 & 33  \\
\midrule
Parallel-Synthesis & 23 & 19 & \textbf{2} & \textbf{36}  \\
\bottomrule
\end{tabular}%
}
\end{minipage}
\hfill
\begin{minipage}[t]{0.52\linewidth}
\centering
\captionof{figure}{TTFT comparison between \textsyn, CacheBlend and \proj. The label indicate speedup to \textsyn.}
\includegraphics[width=\linewidth]{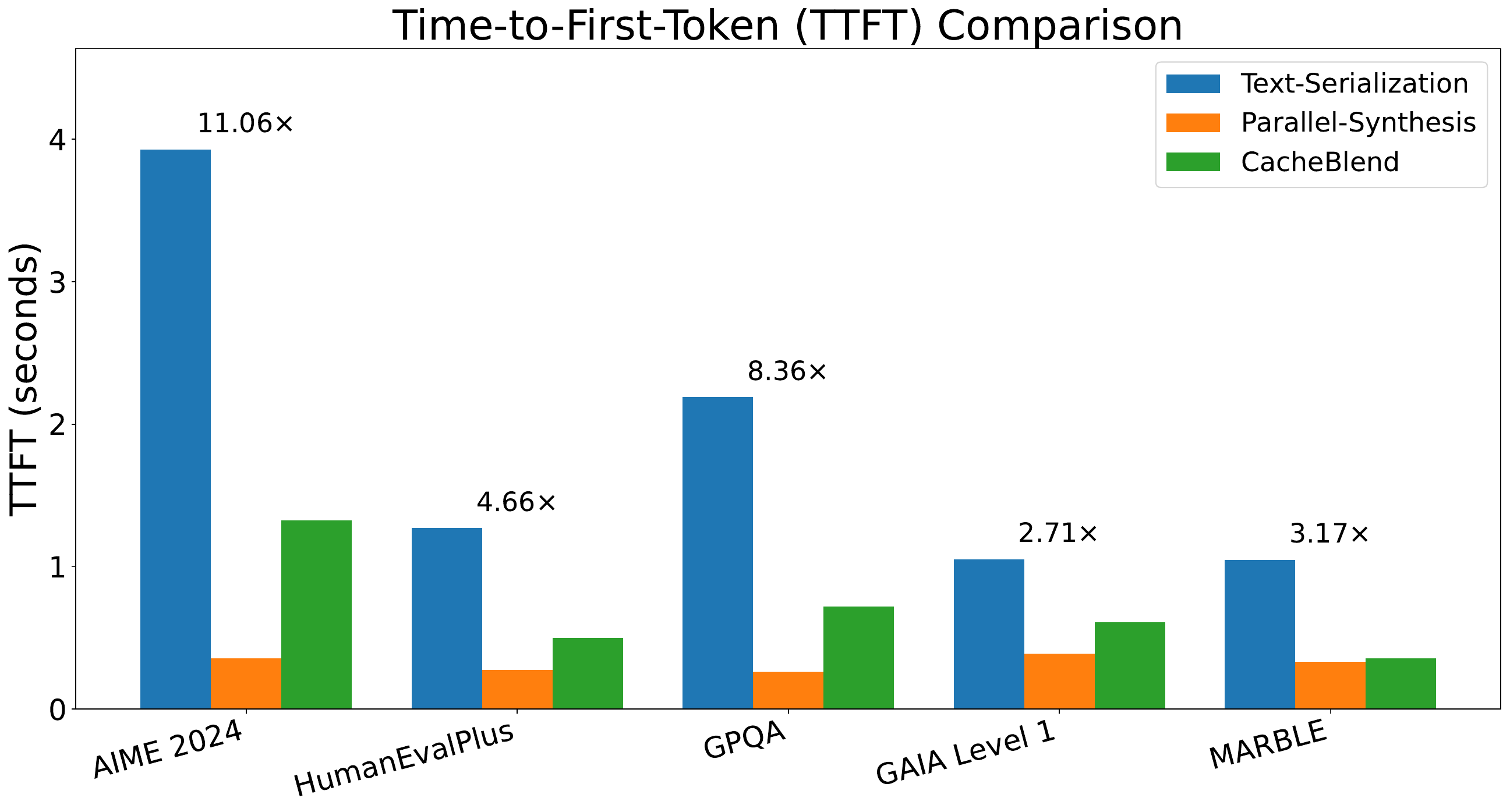}

\label{fig:ttft_comparison}
\end{minipage}
\vspace{-8mm}
\end{figure}

\textbf{RQ1: How does \proj compare to \textsyn in performance and efficiency?}

Based on Tables~\ref{tab:main_problem} and \ref{tab:main_agentic}, \proj matches or surpasses \textsyn on 7 of 9 datasets, with only small gaps on the remaining two. This shows that, after post-training, the synthesizer can reason effectively over prior agents’ outputs without requiring those outputs to be reintroduced as text in the prompt. The gains are pronounced on reasoning-heavy datasets, including math and science benchmarks and the MARBLE database task, where the synthesizer must integrate rich reasoning traces from parallel workers. This suggests that explicit text can be a limited communication interface for reasoning-heavy content, whereas cache sharing may preserve richer latent reasoning states for synthesis. This aligns with prior work on latent-space reasoning~\citep{haotraining, zhengthought, su2025token}. By contrast, explicit text can still be advantageous in more evidence-centric cases, such as GAIA examples involving search results or MBPP-Plus problems with explicit code outputs, where text may provide a clearer and more robust evidence representation.

To quantify the computational savings of \proj from avoiding the re-prefill of parallel worker outputs required by \textsyn, we measure the synthesizer’s Time-to-First-Token (TTFT) on five datasets. For fairness, we include in \proj’s latency both the cache transformation time, covering cache mapping,  positional re-encoding, and the TTFT. Fig.~\ref{fig:ttft_comparison} shows that \proj consistently reduces latency across all datasets, achieving 2.5$\times$--11$\times$ speedups over \textsyn. We also compare with CacheBlend using a 0.15 selective-recomputation ratio, \proj remains an average 2$\times$ faster while achieving higher downstream performance. These gains become larger as worker agent output trajectories grow longer.

\textbf{RQ2: Does \proj perform genuine reasoning-based synthesis over parallel agent trajectories, beyond simple answer aggregation?}

An important question is whether \proj uses parallel agent trajectories as rich synthesis context, rather than treating them as a set of final answers to be aggregated by voting. The \textit{Voting} baseline provides a direct test for this distinction. Across 8 of the 9 datasets, \proj outperforms pure majority voting, indicating that its gains cannot be explained by simply recovering the most common worker answer. MedQA is the only exception, as this multiple-choice task is particularly amenable to majority voting. 

These results suggest that \proj can exploit trajectory-level signals beyond workers' final answers, including intermediate reasoning steps, evidence quality, and disagreements among workers. This capability is especially important for tasks such as GAIA, where the correct answer is not always supported by the majority of workers' final answers. Case studies in Appendix~\ref{app:case_study} show that \proj can ground its synthesis in intermediate evidence from parallel agents; in some examples, it even performs an additional tool call for verification when needed. By contrast, although KVLINK performs reasonably on some datasets, it lags behind \proj overall and often jumps directly to a final answer without reasoning over the worker trajectories or provided instruction. This suggests that post-training for RAG-style cache reuse may not be sufficient for the trajectory-level reasoning and instruction-following ability needed for parallel-agent synthesis.

\textbf{RQ3: Importance of post-training for robust and general parallel cache synthesis.}

Post-training is critical for making direct cache reuse reliable in parallel synthesis. Although \proj is not trained on any of the nine downstream benchmarks, it transfers effectively across all of them, suggesting that the learned synthesis behavior generalizes beyond the training distribution. In contrast, inference-time cache-reuse methods such as APE, CacheBlend, and the no-training variant of \proj perform substantially worse, whereas the post-trained KVLINK baseline is comparatively stronger. This shows that robust parallel cache synthesis requires explicit training-based adaptation, rather than direct inference-time cache calibration alone. Sec.~\ref{sec:ablation} further analyzes the contribution of training data and design choices.

\vspace{-1mm}
\subsection{Ablation Studies}
\label{sec:ablation}
\begin{table}[t]
\vspace{-9mm}
\centering
\caption{Ablation study on checkpoints and model modules. The full default model uses the merged Training Track 1+2 checkpoint together with cache mapping and Synthesizer LoRA.}
\label{tab:ablation_combined}
\resizebox{0.95\linewidth}{!}{%
\begin{tabular}{ll|ccc|cc}
\toprule
Ablation & Variant
& AIME 2024
& HumanEvalPlus
& GPQA
& GAIA Level 1
& MARBLE \\
\midrule
\multirow{4}{*}{\textit{Checkpoint}}
& No training          & 10.00 & 38.41 & 15.15    & 16/53 & 22/100 \\
& Track 1              & 46.67 & 91.46 & 55.56 & 21/53 & 30/100 \\
& Track 2              & 53.33 & 88.41 & 51.52 & 21/53 & 44/100 \\
& Sequential Track 1+2 & 36.67 & 76.22 & 49.49 & 23/53 & 25/100 \\
\midrule
\textit{Module}
& Synthesizer LoRA only & 60.00 & 88.41 & 40.40 & 22/53 & 35/100 \\
\midrule
\multicolumn{2}{l|}{Full merged model checkpoint (default)}
& 63.33 & 90.85 & 52.02 & 23/53 & 36/100 \\
\bottomrule
\end{tabular}%
}
\vspace{-2mm}
\end{table}
\begin{table}[t]
\centering
\caption{Ablation study on the information passed to the synthesizer on GAIA Level 1. Correctness reports the number of correctly answered samples out of 53 samples. TTFT is measured in seconds.}
\label{tab:gaia_info_ttft_ablation}
\resizebox{0.9\linewidth}{!}{%
\begin{tabular}{l|cc|cc}
\toprule
\multirow{2}{*}{Information for synthesizer}
& \multicolumn{2}{c|}{Correctness}
& \multicolumn{2}{c}{TTFT (s)} \\
& \proj
& \textsyn
& \proj
& \textsyn \\
\midrule
Final model output     & 23 & 24 & 0.3769 & 1.0510 \\
Each-turn model output & 18 & 22 & 0.3788 & 1.5053 \\
Full trajectory      & 26 & 26 & 0.6847 & 3.5383 \\
\bottomrule
\end{tabular}%
}
\vspace{-6mm}
\end{table}

\textbf{Ablation on different post-training checkpoints.} We compare four variants on one representative dataset from each task category: no post-training, Track~1 only, Track~2 only, and sequential tuning where Track~2 is applied after Track~1. As shown in Table~\ref{tab:ablation_combined}, Track~1 tends to perform better on simpler single-turn problem-solving tasks, such as HumanEvalPlus and GPQA, whereas Track~2 is stronger on reasoning-heavy math and more complex agentic tasks. This pattern is consistent with the data composition and intended capabilities of the two training tracks described in Sec.~\ref{sec:training}. The final merged checkpoint achieves the best overall balance, while sequential tuning typically underperforms both single-track checkpoints and the merged model. This suggests that sequential tuning fails to preserve the complementary strengths learned from the two tracks.

\textbf{Ablation on framework modules.} We further examine the contributions of the learnable cache mapper and synthesizer LoRA. Since the no-training variant and inference-time cache-reuse baseline APE perform poorly, synthesizer adaptation is necessary for parallel cache synthesis. Table~\ref{tab:ablation_combined} further shows that adding the learnable cache mapper consistently improves over the synthesizer-LoRA-only variant, indicating that cache mapping provides complementary benefits beyond LoRA adaptation.

\textbf{Ablation on communicated trajectory information.}
In single-turn tasks, each worker produces a single response, so the communicated context is unambiguous. In multi-turn ReAct-style tasks, however, a worker trajectory contains instructions, intermediate model responses, tool calls, and tool observations, raising the question of which parts should be passed to the synthesizer. Table~\ref{tab:gaia_info_ttft_ablation} compares three choices on GAIA Level 1: \textit{final output}, which passes only the worker's final response; \textit{each-turn output}, which passes all model-generated responses but excludes instructions and tool observations; and \textit{full trajectory}, which passes the complete worker history. Full trajectories achieve the highest accuracy by preserving tool observations, but incur the largest latency, especially for Text-Serialization. Final outputs provide the best accuracy-efficiency tradeoff and are used as our default. Each-turn outputs perform worse, likely because removing tool observations leaves fragmented, weakly grounded traces. These results show that \proj can synthesize from different levels of trajectory information while benefiting most from concise, coherent worker outputs.


\vspace{-1mm}
\section{More Related Works}


\textbf{LLM-based Multi-agent Systems (MAS).}
Recent LLM-based multi-agent systems increasingly move beyond sequential execution~\citep{yao2022react, zhang2024chain} by organizing agent computation as structured workflows, often in the form of computational graphs or DAGs, where LLM/tool nodes correspond to intermediate subtasks and edges specify execution dependencies~\citep{zhuge2024gptswarm,zhangaflow, yu2025dyntaskmas,qin2025flash}. Within these graph-structured workflows, the parallel-then-synthesize pattern appears in several forms. First, some workflows parallelize solution and rationale generation, where multiple agents or model calls independently solve the same problem and their candidate answers are later aggregated through voting, debate, ranking, or response fusion~\citep{du2024improving, li2024more, jiang2023llm, wangmixture}. Second, some workflows parallelize tool use and open-ended exploration, where independent retrieval/API calls or tool-using subagents are executed along different search directions before a lead agent synthesizes their observations~\citep{kim2024llm,anthropic2025multiagentresearch,wu2025gap,miloradovic2023optimizing, zhang2025agentorchestra, chai2025scimaster}. Lastly, some works parallelize over input contexts: long documents or retrieved evidence are split into chunks, processed by separate agents and then aggregated into a final answer~\citep{zhao2024longagent,zhou2025llm,liu2025scaling}. 

\textbf{Latent communication in multi-agent systems.}
Many work studies how LLM agents can communicate beyond decoded texts. They mainly fall into two lines: one uses latent representations as compact thought-level messages for direct agent communication~\citep{zhengthought,zou2025latent,du2025enabling}, while the other treats KV caches as transferable execution states, enabling agents or models to reuse, share, or fuse internal context without decoding it into text~\citep{yekvcomm,shi2025kvcomm,fu2025cache}. These works mainly study pairwise or sequential inter-agent transfer, where one agent’s internal state is passed to another agent. In contrast, our setting targets the many-to-one convergence pattern in non-sequential workflows: several independently generated branch caches must be jointly interpreted by a downstream synthesizer. A recent work~\citep{jin2026agent} investigates reusable latent primitives for multi-agent collaboration, including parallel branches voting and selection, whereas we focus on training a general parallel-agent synthesizer.



\vspace{-1mm}
\section{Conclusion and Future Direction}
We introduced \proj, a plug-and-play framework that enables a downstream synthesizer to directly consume the KV caches produced by parallel worker agents. Through post-training, \proj performs effectively across a diverse set of parallel-then-synthesize tasks, matching or surpassing text-based synthesis while substantially reducing redundant computation. \proj is not yet ready to replace text-based communication in all agent frameworks, especially for highly complex structured workflows. Nevertheless, our results provide encouraging evidence that direct cache-based synthesis is a promising interface for parallel agent branches. If this direction succeeds, DAG-structured workflows could be executed and merged more natively, without the repeated computational overhead of text serialization and context refilling. Achieving this goal will likely require scaling post-training to broader and larger collections of agent workflow data, covering more diverse tool-use patterns, branching structures, and synthesis objectives.

\section*{Acknowledgments}

S. Liu, M. Li, H. Wang, and P. Li are partially supported by the NSF under awards IIS-2239565, CCF-2402816, IIS-2435957; the Meta Grant; the NVIDIA Academic Grant Program; and the IDEaS Cyberinfrastructure Awards.

\clearpage
\newpage
\bibliographystyle{assets/plainnat}
\bibliography{paper}

\clearpage
\newpage
\beginappendix

\section{Technical appendices and supplementary material}
We provide additional details on the datasets used for post-training and downstream evaluation in Appendix~\ref{app:data}. Appendix~\ref{app:training_config} reports the experimental configurations, and Appendix~\ref{app:prompts} specifies the prompts used for each dataset. Finally, Appendix~\ref{app:case_study} presents case studies that qualitatively illustrate the reasoning capability of the trained \proj when synthesizing information from parallel agent trajectories.

\subsection{Datasets Details}
\label{app:data}
This section will include omitted detail of the dataset introduction and some statistics of the post-training datasets as well as the downstream evaluation datasets.
\subsubsection{Post-training datasets}
The statistics for the post-training datasets are included in Table~\ref{tab:post-train-data}. 
\begin{itemize}
    \item Wildchat~\citep{zhao2024wildchat}: A large-scale corpus of real-world user–ChatGPT conversations collected from opt-in users. It contains diverse user intents, languages, and multi-turn interaction patterns, making it useful for studying realistic dialogue histories.
    \item LMSYS-Chat~\citep{zheng2023lmsyschat1m}: A real-world LLM conversation dataset collected from the Vicuna demo and Chatbot Arena. It contains conversations with multiple state-of-the-art LLMs and is useful for modeling natural user–assistant interactions.
    \item UltraChat~\citep{ding2023enhancing}: A large-scale synthetic multi-turn instructional dialogue dataset generated with ChatGPT-style user and assistant roles. It covers broad open-domain instruction-following scenarios and is designed for chat model training.
    \item Toucan-single-turn~\citep{xu2025toucan}: A subset of Toucan-1.5M, a large tool-agentic dataset synthesized from real-world MCP environments. Here, we filter out only single-turn interaction setting but with parallel toolcalls. We train the model to generate corresponding response given the parallel tool-calling execution.
    \item Toucan-multi-turn~\citep{xu2025toucan}: The multi-turn subset of Toucan, containing tool-agent trajectories extended to multi-turn interactions. It is useful for training models to plan tool use conditioned on previous user–agent interactions.
    \item DTA-Tool~\citep{zhu2025dividethenaggregateefficienttoollearning}: A parallel tool invocation dataset built for DTA-Llama. It converts traditional tool-search trajectories into DAG-style parallel tool invocation traces, where the model learns to divide a task into parallel tool-using subtasks and aggregate the results.
    \item FLAN~\citep{weifinetuned}: An instruction-tuning collection that formats many NLP tasks with natural language instructions, including zero-shot, few-shot, and chain-of-thought templates. In your setting, it is suitable for constructing current-question/current-answer pairs with previous in-context examples.
    \item 2WikiMultiHopQA~\citep{xanh2020_2wikimultihop}: A multi-hop question answering dataset built from structured Wikidata and unstructured Wikipedia evidence. It provides explicit evidence and reasoning paths, making it useful for training or evaluating context-grounded multi-hop reasoning.
\end{itemize}
\label{app:post-train-data}
\begin{table}[t]
\centering
\caption{Statistics of training datasets.}
\label{tab:post-train-data}
\resizebox{\linewidth}{!}{%
\begin{tabular}{lrrrr}
\toprule
Dataset 
& Samples 
& Avg $m$ 
& Avg length of $z_j$ 
& Avg supervised tokens/sample \\
\midrule
\texttt{wildchat}               & 206,788 & 4.77 & 471.79 & 343.64 \\
\texttt{ultrachat}              & 140,520 & 2.72 & 380.39 & 284.02 \\
\texttt{lmsys\_chat}            & 175,635 & 4.42 & 234.08 & 267.59 \\
\texttt{toucan\_single\_turn} & 141,713 & 2.67 & 880.03 & 400.50 \\
\texttt{toucan\_multi\_turn}  & 1,171   & 2.49 & 826.69 & 57.37  \\
\texttt{dta\_tool}              & 14,459  & 2.41 & 288.99 & 231.10 \\
\texttt{flan}                   & 33,351  & 4.16 & 95.96  & 27.63  \\
\texttt{2wiki\_multihopqa}      & 167,254 & 9.93 & 82.76  & 133.55 \\
\texttt{Browsecomp\_text\_traj}      & 1211 & 3 & 858.33  & 735.07 \\
\bottomrule
\end{tabular}%
}
\end{table}

\subsubsection{Evaluation datasets}
\label{app:eval-data}
\textbf{Math Problem Reasoning}
\begin{itemize}
    \item AIME~\citep{maxwell_jia_aime_2024, aime25}: AIME 2024 and 2025 each contains 30 competition-level problems from the American Invitational Mathematics Examination 2024 and 2025. The questions are  span algebra, geometry, number theory,
    and combinatorics, and require precise numeric answers with typically 1–3 digits and are intended for evaluating LLM mathematical reasoning on challenging high-school olympiad-style questions.
    \item GSM8K~\citep{cobbe2021gsm8k}: It consists of 8.5K human-written grade-school math word problems, split into 7.5K training and 1K test examples. Each problem typically requires two to eight reasoning steps and a sequence of elementary arithmetic operations, making it a standard benchmark for evaluating multi-step mathematical reasoning rather than isolated calculation.
\end{itemize}

\textbf{Code Generation}
\begin{itemize}
    \item HumanEval-Plus~\citep{liu2023your}: It is a Python code-generation benchmark for evaluating the functional correctness of LLM-synthesized programs via execution against much more exhaustive unit tests than typical coding datasets. It pairs HumanEval-style programming prompts (function signature + docstring) with automatically generated, type-aware, adversarially diverse test inputs to better catch subtle logical errors that slip past sparse test suites, making it a stricter measure of whether models produce semantically correct, runnable code rather than merely plausible-looking solutions.
    \item MBPP-Plus~\citep{austin2021program, liu2023your}: It extends the Mostly Basic Python Problems benchmark, which contains short Python programming tasks designed to be solvable by entry-level programmers. Compared with the original MBPP tasks, which include only a small number of tests, MBPP-Plus augments the benchmark with substantially richer unit tests, making it a more rigorous evaluation of functional correctness, edge-case handling, and the robustness of generated Python code.
\end{itemize}

\textbf{Scientific QA}
\begin{itemize}
    \item GPQA-Diamond~\citep{rein2024gpqa}: It is the hardest subset of the GPQA benchmark, consisting of 198 graduate-level multiple-choice questions written by domain experts in physics, biology, and chemistry. It is designed to be “Google-proof,” with carefully engineered distractors that require deep conceptual understanding, cross-topic reasoning, and multi-step scientific inference rather than surface pattern matching.
    \item MedQA~\citep{yang2025llm}: It comprises real medical licensing–style multiple-choice questions that evaluate biomedical knowledge and clinically grounded reasoning. Solving items often requires integrating the problem narrative with domain expertise to make diagnostic or management decisions under plausible distractors, making it a practical proxy for professional-level medical QA performance.
\end{itemize}

\textbf{GAIA}~\citep{mialon2023gaia}: A benchmark for general AI assistants that tests whether a system can solve real-world, conceptually simple (for humans) questions that nonetheless require multi-step reasoning, web browsing, multimodal understanding, and tool use (e.g., reading files, searching, computing). In our setting, we use 5 max toolcalls for each parallel worker. 

\textbf{MARBLE Database}~\citep{zhu2025multiagentbench}: It is a multi-agent database diagnosis task from MultiAgentBench/MARBLE, where a PostgreSQL database exhibits performance anomalies and agents must collaboratively identify the underlying root cause. Solving the task requires agents to inspect database system views such as query statistics and lock information, communicate across multiple rounds, and synthesize evidence across possible causes including large data fetching or insertion, lock contention, redundant indexes, and vacuum-related performance issues. In our setting, we mainly let each parallel worker agent investigate one root cause and pass its analysis to the synthesizer and let the synthesizer to decide the final root causes.

\subsection{Experimental Configuration}
\label{app:training_config}

\textbf{Post-Training.} During post-training, for both tracks, we jointly optimize the cache affine mapper and a judger LoRA adapter. The LoRA adapter uses rank 16, alpha 32, and dropout 0.0, and is applied to the attention projection modules \texttt{q\_proj}, \texttt{k\_proj}, \texttt{v\_proj}, and \texttt{o\_proj}. We optimize the trainable modules with AdamW using a learning rate of $1\times10^{-4}$ and weight decay 0.0. We use 5 A100 for training with GPU memory 80G. Post-training for track 1 takes several days and for track 2 takes hours less than a day.

\textbf{Evaluation.}
Note that for the GAIA dataset, we use the open validation set as our downstream performance evaluation. This is because there is no visible answer for the test set. And using GAIA validation for evaluation is a common practice in academic papers~\citep{nguyendynasaur, soni2026coding, zhang2025optimizing}. 

For all datasets inference, we set the max new tokens for the parallel worker and the synthesizer agent to 4096 by default except the AIME datasets we set to 6000. We use 5 max tool steps for GAIA and MARBLE dataset. Regarding baselines, for the APE baseline, we use the default scale and temperature in their code as both 0.9. For the CacheBlend, we use 0.15 recomputation rate.

\subsection{Prompts}
\label{app:prompts}
Below we show the prompts we use for \proj for each dataset. The \textsyn prompt follow the same instruction except we add the explicit past agent outputs into the prompt. 

\begin{promptbox}{Parallel-Synthesis Prompts on Math Problem Solving (GSM8K / AIME2024 / AIME2025)}

\textbf{System Prompt for All Agents:}

You are Qwen, created by Alibaba Cloud. You are a helpful assistant.

\vspace{0.5em}
\textbf{Prompt for Parallel Agent:}

Solve the input question step by step and put the final answer inside
\verb|\boxed{YOUR_FINAL_ANSWER}|. The final answer must be a single numerical
value with no extra text or units inside the box.

\textbf{Input Question:}

\texttt{\{question\}}

\textbf{Your response:}

\vspace{0.8em}
\textbf{Prompt for Synthesizer Agent:}

You are a task summarizer. Solve the input question step by step and put the
final answer inside \verb|\boxed{YOUR_FINAL_ANSWER}|. The final answer must be
a single numerical value with no extra text or units inside the box. You are
provided with latent information from previous agents as reference.

\textbf{Input Question:}

\texttt{\{question\}}

\textbf{Your response:}

\end{promptbox}

\paragraph{Code-generation prompts.}
HumanEval-Plus and MBPP-Plus use the same prompt template for code generation.

\begin{promptbox}{Parallel-Synthesis Prompts on Code-Generation Tasks (HumanEval-Plus / MBPP-Plus)}

\textbf{System Prompt for All Agents:}

You are Qwen, created by Alibaba Cloud. You are a helpful assistant.

\vspace{0.5em}
\textbf{Prompt for Parallel Agent:}

Write a correct and self-contained Python solution for the input question.
You must put all python code in a markdown code block. For example:

\begin{quote}
\ttfamily\small
\verb|```python|\\
import math\\
def add(a, b):\\
\hspace*{1em}return a + b\\
\verb|```|
\end{quote}

Do not add any other contents inside the markdown code block.

\textbf{Input Question:}

\texttt{\{question\}}

\textbf{Your response:}

\vspace{0.8em}
\textbf{Prompt for Synthesizer Agent:}

You are a task summarizer. Write a correct and self-contained Python solution
for the input question. You must put all python code in a markdown code block.
For example:

\begin{quote}
\ttfamily\small
\verb|```python|\\
import math\\
def add(a, b):\\
\hspace*{1em}return a + b\\
\verb|```|
\end{quote}

Do not add any other contents inside the markdown code block. You are provided
with latent information from previous agents as reference.

\textbf{Input Question:}

\texttt{\{question\}}

\textbf{Your response:}

\end{promptbox}

\begin{promptbox}{Parallel-Synthesis Prompts on Scientific QA Tasks (GPQA / MedQA)}

\textbf{System Prompt for All Agents:}

You are Qwen, created by Alibaba Cloud. You are a helpful assistant.

\vspace{0.5em}
\textbf{Prompt for Parallel Agent:}

Solve the input question step by step and put the final answer inside
\verb|\boxed{YOUR_FINAL_ANSWER}|. Your final answer must be selected from
A, B, C, D. For example, \verb|\boxed{A}|. Do not add any other contents
inside the box.

\textbf{Input Question:}

\texttt{\{question\}}

\textbf{Your response:}

\vspace{0.8em}
\textbf{Prompt for Synthesizer Agent:}

You are a task summarizer. Solve the input question step by step and put the
final answer inside \verb|\boxed{YOUR_FINAL_ANSWER}|. Your final answer must
be selected from A, B, C, D. For example, \verb|\boxed{A}|. Do not add any
other contents inside the box. You are provided with latent information from
previous agents as reference.

\textbf{Input Question:}

\texttt{\{question\}}

\textbf{Your response:}

\end{promptbox}

\paragraph{GAIA tool-use prompts.}
GAIA uses a tool-use prompt with web search, webpage visiting, Python execution, and file parsing tools. The attached-file field is omitted when a sample has no files.

\begin{promptbox}{Parallel Agent Prompt for GAIA}

\textbf{System Prompt:}

You are a deep research assistant with tool-use capability.

\textbf{Current date:} \texttt{\{current\_date\}}

You may call one or more tools to solve the user query. Tool signatures are listed below:

\begin{itemize}
    \item \texttt{search(query)}: Perform web search.
    \item \texttt{visit(url, goal)}: Visit webpage(s) and return extracted text relevant to a goal.
    \item \texttt{PythonInterpreter(code)}: Execute Python code. Any output you want must be printed.
    \item \texttt{parse\_file(files)}: Parse local files and return readable text.
\end{itemize}

When you need a tool, output exactly one tool call using XML tags:
\texttt{\textless tool\_call\textgreater ... \textless/tool\_call\textgreater}.

If the tool is \texttt{PythonInterpreter}, put code right after the JSON in
\texttt{\textless code\textgreater...\textless/code\textgreater} tags.

Tool outputs will be returned in:
\texttt{\textless tool\_response\textgreater ... \textless/tool\_response\textgreater}.

When you are ready to finish, provide your final answer inside:
\texttt{\textless answer\textgreater...\textless/answer\textgreater}.

Never fabricate tool outputs.

\vspace{0.8em}
\textbf{User Prompt:}

\textbf{Target Question:} \texttt{\{question\}}

You are a helpful assistant. You must reason step-by-step to solve the provided
Target Question. Return the final answer only inside
\texttt{\textless answer\textgreater YOUR\_FINAL\_ANSWER \textless/answer\textgreater}.

\textbf{Attached files you can parse with \texttt{parse\_file}:}

\texttt{\{attached\_file\_path\}}

\end{promptbox}

\begin{promptbox}{Synthesizer Prompt for GAIA}

\textbf{System Prompt:}

You are a deep research assistant with tool-use capability.

\textbf{Current date:} \texttt{\{current\_date\}}

You may call one or more tools to solve the user query. Tool signatures are listed below:

\begin{itemize}
    \item \texttt{search(query)}: Perform web search.
    \item \texttt{visit(url, goal)}: Visit webpage(s) and return extracted text relevant to a goal.
    \item \texttt{PythonInterpreter(code)}: Execute Python code. Any output you want must be printed.
    \item \texttt{parse\_file(files)}: Parse local files and return readable text.
\end{itemize}

When you need a tool, output exactly one tool call using XML tags:
\texttt{\textless tool\_call\textgreater ... \textless/tool\_call\textgreater}.

If the tool is \texttt{PythonInterpreter}, put code right after the JSON in
\texttt{\textless code\textgreater...\textless/code\textgreater} tags.

Tool outputs will be returned in:
\texttt{\textless tool\_response\textgreater ... \textless/tool\_response\textgreater}.

When you are ready to finish, provide your final answer inside:
\texttt{\textless answer\textgreater...\textless/answer\textgreater}.

Never fabricate tool outputs.

\vspace{0.8em}
\textbf{User Prompt:}

You are a Task Summarizer.

Use the previous agents' contexts as reference. They may be provided as explicit
text or through KV memory.

Reason over their answers and evidence step by step, compare the candidate
answers, and give the best possible final answer. If two or more agents give
the same plausible answer, use it directly. If only one agent gives a plausible
answer, consider whether its evidence fits the question instead of dismissing
it just because it is a minority answer.

Use tools only when the previous agents do not give any plausible direct
answer, or when all candidate answers are clearly unsupported. If you use a
tool, use only the listed tools, make one targeted check, and then answer.

\textbf{Content from Previous Agents:}

\texttt{[provided through KV memory]}

\textbf{Input Question:}

\texttt{\{question\}}

\textbf{Attached files, use \texttt{parse\_file} if needed:}

\texttt{\{attached\_file\_path\}}

\textbf{Guidelines:}

\begin{itemize}
    \item Prefer synthesizing prior agents over starting from scratch.
    \item If agents disagree, choose the best-supported answer, not necessarily the majority answer.
    \item If agents agree on a plausible answer, do not verify it with tools.
    \item If uncertain, still choose the most plausible answer from the prior agents' work.
    \item Prefer a direct short candidate answer over refusals, tool traces, or generic explanations.
    \item Do not output an empty response, markdown fence, tool schema, raw XML, or raw tool response.
    \item Output only the final answer inside
    \texttt{\textless answer\textgreater YOUR\_FINAL\_ANSWER \textless/answer\textgreater}.
\end{itemize}

\textbf{Your response:}

\end{promptbox}

\paragraph{MARBLE Database prompts.}
MARBLE Database uses role-conditioned database diagnosis agents and a final diagnosis synthesizer. Each worker is assigned a role profile and a database diagnosis subtask. The \texttt{\{agent\_profile\}} and  \texttt{\{assigned\_task\}}includes the task assigned to that agent like "agent1 will explore the possibility of INSERT\_LARGE\_DATA as a root cause. Recommended tables: `pg\_stat\_statements`. You can search for INSERTs."

\begin{promptbox}{Parallel Agent Prompt for MARBLE Database}

\textbf{System Prompt:}

You are ``\texttt{\{agent\_id\}}'': ``\texttt{\{agent\_profile\}}''.

As a role-playing agent, you embody a dynamic character with unique traits,
motivations, and skills. Your goal is to engage not only with users but also
with other agents in the environment. Collaborate, compete, or form alliances
as you navigate through immersive storytelling and challenges. Interact
meaningfully with fellow agents, contributing to the evolving narrative and
responding creatively to their actions. Maintain consistency with your
character's background and personality, and be prepared to adapt to the
evolving dynamics of the scenario.

Remember, your responses should enhance the experience and encourage user
engagement while enriching interactions with other agents.

\vspace{0.5em}
You are a database diagnosis assistant with tool-use capability.

\textbf{Current date:} \texttt{\{current\_date\}}

You may call one or more tools to solve the assigned database diagnosis task.
Tool signatures are listed below:

\begin{itemize}
    \item \texttt{query\_db(sql)}: Query the PostgreSQL database with the given SQL statement.
    Use \texttt{pg\_stat\_statements} and related catalog views. Use one query at a time
    and avoid queries that may hang the database.
\end{itemize}

When you need a tool, output exactly one tool call using XML tags:
\texttt{\textless tool\_call\textgreater ... \textless/tool\_call\textgreater}.

Tool outputs will be returned in:
\texttt{\textless tool\_response\textgreater ... \textless/tool\_response\textgreater}.

When you are ready to finish, answer directly in plain text or JSON.

Never fabricate tool outputs.

\vspace{0.8em}
\textbf{User Prompt:}

\textbf{Agent ID:} \texttt{\{agent\_id\}}

\textbf{Task:} \texttt{\{assigned\_task\}}

Use the available tools if needed, then provide your final response in plain text.
Start your final response with exactly:

\begin{quote}
\ttfamily\small
This is the response from \{agent\_id\}:
\end{quote}

\end{promptbox}

\begin{promptbox}{Synthesizer Prompt for MARBLE Database}

\textbf{User Prompt:}

You are the final database diagnosis judger. You are given the prior agents'
investigations, including their findings, cited evidence, and any reported
tool-use results. Your job is to inspect and summarize those investigations
carefully, and produce the best final diagnosis.

\textbf{Work in this order:}

\begin{enumerate}
    \item Inspect the current task background and the allowed label set.
    \item Inspect what each previous agent investigated and what evidence they found.
    \item Choose the most strongly supported root causes and summarize why.
\end{enumerate}

\textbf{Task:}

\texttt{\{question\}}

\textbf{Current Objective:}

Identify the most likely root cause labels behind the current database
performance issue.

\textbf{Allowed Labels:}

\texttt{'\{LABEL\_1\}', '\{LABEL\_2\}'}

\textbf{Decision Rule:}

Select at most 2 labels, and only choose labels that are supported by the
previous agents' findings.

\textbf{Previous Agent Investigations:}

\texttt{[provided through KV memory]}

Decide which root causes are best supported by the agents' findings. Before the
final answer, write a short Rationale based on the previous agents' diagnoses.
Briefly summarize what the agents found, compare the competing explanations,
and choose the root causes that seem most likely overall.

Use the agents' SQL/tool evidence when it is available, but do not require
perfect evidence for every selected label. If some findings are weak or
conflicting, make the best diagnosis from the strongest available signals
rather than returning no answer.

Do not invent row counts, lock states, SQL queries, tables, indexes, or agent
conclusions that were not in the prior investigations.

Return the final selected root-cause labels inside exactly one
\texttt{\textless answer\textgreater...\textless/answer\textgreater} block.
Inside the answer block, output only a JSON array of labels. The answer array
must contain at most 2 labels. If only one label is well supported, return a
one-element array.

Choose labels only from this list:

\texttt{'\{LABEL\_1\}', '\{LABEL\_2\}'}

Do not put justification inside the \texttt{\textless answer\textgreater} block.
Do not wrap the answer in Markdown fences. Do not output more than one
\texttt{\textless answer\textgreater} block.

\end{promptbox}

\subsection{Cases Studies of \proj}
\label{app:case_study}
\paragraph{Case study: synthesis without majority agreement.}
The following AIME 2024 example illustrates that \proj can synthesize useful
partial reasoning from multiple workers even when none of the workers produces
the correct final answer.

\begin{casebox}{AIME 2024 Case Study: Correct Synthesis without a Majority Answer}

\textbf{Problem.}
Find the number of triples of nonnegative integers $(a,b,c)$ satisfying
$a+b+c=300$ and
\[
a^2b+a^2c+b^2a+b^2c+c^2a+c^2b=6{,}000{,}000.
\]

\textbf{Gold answer:} $\boxed{601}$ \hfill
\textbf{Synthesizer prediction:} $\boxed{601}$

\textbf{No-majority worker outputs.}
The three parallel workers produce three distinct final answers, so there is no
majority answer to vote for.

\begin{center}
\small
\begin{tabularx}{0.98\linewidth}{l c X}
\toprule
\textbf{Source} & \textbf{Final answer} & \textbf{Useful evidence in the trace} \\
\midrule
Sub-agent 1 &
\texttt{2} &
Correctly reduces the original equation to
$100(ab+bc+ca)-abc=2{,}000{,}000$, but does not complete the count. \\
Sub-agent 2 &
\texttt{100} &
Finds the decisive factorization
$(a-100)(b-100)(c-100)=0$, but does not produce the correct final count. \\
Sub-agent 3 &
\texttt{200} &
Checks valid examples such as $(100,100,100)$ and $(200,100,0)$, providing
sanity checks for the reduced equation. \\
\bottomrule
\end{tabularx}
\end{center}

\begin{evidencebox}{Complementary evidence extracted from workers}

\textbf{Sub-agent 1: equation reduction.}
Using $a+b+c=300$, the worker derives
\[
300(ab+bc+ca)-3abc=6{,}000{,}000,
\]
and hence
\[
100(ab+bc+ca)-abc=2{,}000{,}000.
\]

\textbf{Sub-agent 2: key factorization.}
The worker observes that
\[
(a-100)(b-100)(c-100)
=abc-100(ab+bc+ca)+100^2(a+b+c)-100^3.
\]
Since $a+b+c=300$ and $100(ab+bc+ca)-abc=2{,}000{,}000$, this becomes
\[
(a-100)(b-100)(c-100)=0.
\]

\textbf{Sub-agent 3: sanity checks.}
The worker verifies examples such as $(200,100,0)$ and $(100,100,100)$,
both of which satisfy the reduced equation
$100(ab+bc+ca)-abc=2{,}000{,}000$.

\end{evidencebox}

\begin{evidencebox}{Synthesizer reconstruction}

The synthesizer combines the partial evidence rather than voting over final
answers. It first identifies the critical structural evidence from Sub-agent 2,
as shown by its opening reasoning:

\begin{quote}
\small\itshape
``From my earlier analysis, I found that the equation simplifies to
$(a - 100)(b - 100)(c - 100) = 0$. This means that at least one of the
variables $a$, $b$, or $c$ must be equal to $100$.''
\end{quote}

Although Sub-agent 2 does not produce the correct final count, this
factorization correctly characterizes the solution set. The synthesizer then
uses this evidence to complete the counting argument. Let
$A=\{a=100\}$, $B=\{b=100\}$, and $C=\{c=100\}$. If $a=100$, then
$b+c=200$, giving $201$ nonnegative solutions. Therefore
$|A|=|B|=|C|=201$. For any pair, e.g., $a=b=100$, the remaining variable is
forced to be $100$, so
\[
|A\cap B|=|A\cap C|=|B\cap C|=1,
\qquad
|A\cap B\cap C|=1.
\]
By inclusion-exclusion,
\[
|A\cup B\cup C|
=3\cdot 201-3\cdot 1+1
=601.
\]

Thus, by using the correct intermediate evidence from a worker with an
incorrect final answer, the synthesizer obtains the correct final answer:
\[
\boxed{601}.
\]

\end{evidencebox}

\textbf{Takeaway.}
This example shows that \proj can solve cases where parallel agents do not
agree on a final answer and none of the worker final answers is correct. The
synthesizer succeeds by integrating complementary latent reasoning signals:
one worker provides the reduced symmetric equation, another provides the key
factorization, and a third supplies useful validity checks.

\end{casebox}

\paragraph{Case study: evidence-aware synthesis without tool verification.}
The following GAIA example illustrates that \proj can use prior-agent evidence
to produce a concise correct answer even when no parallel agent gives an exact
final answer.

\begin{casebox}{GAIA Case Study: Synthesizing Prior Tool Evidence without Additional Tool Calls}

\textbf{Question.}
Examine the video at \texttt{https://www.youtube.com/watch?v=1htKBjuUWec}.
What does Teal'c say in response to the question ``Isn't that hot?''

\textbf{Gold answer:} \texttt{Extremely} \hfill
\textbf{Synthesizer prediction:} \texttt{Extremely} \hfill
\textbf{Synthesizer tool calls:} 0

\textbf{No exact-answer majority among workers.}
Although one worker finds the right phrase, none of the worker final answers is
an exact match to the gold answer. The synthesizer normalizes the evidence into
the concise expected answer.

\begin{center}
\small
\begin{tabularx}{0.98\linewidth}{l X c c}
\toprule
\textbf{Source} & \textbf{Final answer} & \textbf{Exact match?} & \textbf{Tool calls} \\
\midrule
Sub-agent 1 &
\texttt{Teal'c responds with, "Hot? No. But I can be."} &
No & 0 \\
Sub-agent 2 &
\texttt{Teal'c responds to the question "Isn't that hot?" with "Extremely."} &
No & 2 \\
Sub-agent 3 &
\texttt{Indeed} &
No & 1 \\
Synthesizer &
\texttt{Extremely} &
Yes & 0 \\
\bottomrule
\end{tabularx}
\end{center}

\begin{evidencebox}{Parallel-agent evidence}

\textbf{Sub-agent 2: retrieved evidence.}
Sub-agent 2 performs a targeted web search for the relevant quote:

\begin{quote}
\small\itshape
``Teal'c response to 'Isn't that hot?' in Stargate SG-1 video''
\end{quote}

The returned preview contains the key evidence:

\begin{quote}
\small\itshape
``Teal'c? Ah! - Isn't that hot?- Extremely. - Stargate SG-1 | Clip.Cafe''
\end{quote}

Its final answer is not an exact match because it includes extra explanatory
text, but the trace contains the correct phrase \texttt{Extremely}.

\textbf{Other agents: incomplete or noisy evidence.}
Sub-agent 1 attempts to answer the video question but gives an incorrect quote.
Sub-agent 3 gives the incorrect final answer \texttt{Indeed}, but the
synthesizer trace indicates that its prior reasoning also mentioned
\texttt{Extremely} as a plausible answer.

\end{evidencebox}

\begin{evidencebox}{Synthesizer reconstruction}

The synthesizer does not call any tools. Instead, it explicitly compares the
previous agents' attempts and identifies which evidence should be trusted:

\begin{quote}
\small\itshape
``The first agent tried to visit the YouTube link but couldn't get the
transcript. The second agent did a web search and found a clip where Teal'c
says `Extremely' to that question. The third agent also mentioned `Extremely'
based on prior knowledge of Stargate quotes.''
\end{quote}

It then reasons over the provenance of the evidence rather than simply voting
over final strings:

\begin{quote}
\small\itshape
``So, both the web search and prior knowledge point to `Extremely' as the
answer. The first agent's attempt to visit the link didn't yield the transcript,
but the other methods confirmed the answer. Therefore, the most plausible answer
is `Extremely.'''
\end{quote}

Finally, the synthesizer outputs the normalized answer:

\[
\texttt{<answer>Extremely</answer>}.
\]

\end{evidencebox}

\textbf{Takeaway.}
This example shows that \proj is not merely aggregating worker final answers.
The synthesizer reads the provenance of prior attempts, recognizes that one
worker's web search retrieved the relevant quote, downweights a failed direct
video-retrieval attempt, and outputs the concise answer required by the
evaluation format without making any additional tool calls.

\end{casebox}

\paragraph{Case study: conflict- and insufficiency-conditioned tool-use attempts.}
The following GAIA examples illustrate that the synthesizer can recognize when
prior-agent evidence is not sufficient and invoke tools for additional
verification. These are not successful verification cases; rather, they show
that the synthesizer makes the appropriate decision to seek external evidence
when worker outputs are conflicting or incomplete.

\begin{casebox}{GAIA Case Studies: Tool Use When Prior-Agent Evidence Is Insufficient}

\textbf{Primary example: disagreement on an image-based question.}
In one example, the question requires inspecting an attached sheet-music image
to compute an age from bass-clef note letters. The three prior agents produce
mutually inconsistent answers:

\begin{center}
\small
\begin{tabularx}{0.65\linewidth}{l c}
\toprule
\textbf{Source} & \textbf{Final answer} \\
\midrule
Sub-agent 1 & \texttt{3} \\
Sub-agent 2 & \texttt{5} \\
Sub-agent 3 & \texttt{6} \\
\bottomrule
\end{tabularx}
\end{center}

\begin{evidencebox}{Conflict-conditioned file verification}

Because the prior answers conflict and the task depends on an attached image,
the synthesizer attempts to inspect the image with the file-parsing tool:

\begin{quote}
\small\ttfamily
parse\_file(\{``files'': [``.../8f80e01c-1296-4371-9486-bb3d68651a60.png'']\})
\end{quote}

This is the appropriate verification step: the correct answer cannot be
determined reliably from the inconsistent worker outputs alone, and resolving
the disagreement requires reading the attached image.

The verification attempt is blocked because the file is unavailable in the
evaluation environment:

\begin{quote}
\small\ttfamily
[parse\_file] missing: .../8f80e01c-1296-4371-9486-bb3d68651a60.png
\end{quote}

\end{evidencebox}

\textbf{Secondary example: missing information from all prior agents.}
In another example, the question asks for a numerical statistic from
ScienceDirect. All three prior agents fail to retrieve the requested data:

\begin{center}
\small
\begin{tabularx}{0.95\linewidth}{l X}
\toprule
\textbf{Source} & \textbf{Final answer} \\
\midrule
Sub-agent 1 & \texttt{The requested information could not be retrieved...} \\
Sub-agent 2 & \texttt{The requested data could not be retrieved...} \\
Sub-agent 3 & \texttt{Insufficient data available...} \\
\bottomrule
\end{tabularx}
\end{center}

\begin{evidencebox}{Insufficiency-conditioned web verification}

Because the prior agents do not provide any usable answer, the synthesizer
turns to external retrieval. It issues web-search and webpage-visit tool calls:

\begin{quote}
\small\ttfamily
search(``ScienceDirect reference works Life Science Health Sciences 2022 dataset'')
\end{quote}

\begin{quote}
\small\ttfamily
visit(``https://www.sciencedirect.com/science/reference-work'')
\end{quote}

These are the right kinds of actions for the situation: when all prior-agent
evidence is missing or inconclusive, the synthesizer does not guess, but
instead attempts to retrieve the missing information from the source.

The retrieval attempt does not succeed because the visited page is inaccessible:

\begin{quote}
\small\ttfamily
Title: Page not found | ScienceDirect \\
Warning: Target URL returned error 404: Not Found
\end{quote}

\end{evidencebox}

\textbf{Takeaway.}
These two cases show that the synthesizer is not limited to passively combining
prior final answers. Under disagreement (Row 50), it attempts file-based
verification of the relevant artifact; under complete insufficiency of prior
evidence (Row 10), it attempts external web retrieval. Although these
particular cases do not end in successful verification because the relevant file
or webpage is unavailable, they demonstrate that the synthesizer can condition
tool-use decisions on the state of the prior-agent evidence.

\end{casebox}

\paragraph{Case study: evidence-grounded diagnosis in MARBLE Database.}
The following MARBLE Database example illustrates that \proj can synthesize
previous agents' diagnostic analyses, refer to their evidence, and produce a
reasonable final root-cause diagnosis.

\begin{casebox}{MARBLE Database Case Study: Evidence-Grounded Root-Cause Diagnosis}

\textbf{Task.}
The database is used in a file-sharing system with users, files, shared files,
and file-access logs. The model must diagnose the root cause of a database
performance issue from a fixed label set.

\textbf{Gold root cause:} \texttt{INSERT\_LARGE\_DATA} \hfill
\textbf{Judger prediction:} \texttt{["INSERT\_LARGE\_DATA"]}

\begin{evidencebox}{Previous-agent diagnostic evidence}

The previous agents investigate different candidate root causes and provide
both positive and negative evidence.

\begin{center}
\small
\begin{tabularx}{0.98\linewidth}{l l X}
\toprule
\textbf{Agent} & \textbf{Candidate} & \textbf{Evidence from prior analysis} \\
\midrule
Agent 1 &
\texttt{INSERT\_LARGE\_DATA} &
Finds the main positive signal: a query in \texttt{pg\_stat\_statements}
inserting $48{,}000$ rows in one operation. \\
Agent 2 &
\texttt{LOCK\_CONTENTION} &
Checks for non-granted locks, which would indicate waiting transactions, but
finds no evidence of active lock contention. \\
Agent 3 &
\texttt{VACUUM} &
Searches for vacuum-related activity and finds no strong evidence that vacuum
is the source of the performance issue. \\
Agent 4 &
\texttt{REDUNDANT\_INDEX} &
Finds some zero-scan indexes, but notes that many appear to be primary-key or
unique indexes, which are not necessarily redundant. \\
Agent 5 &
\texttt{FETCH\_LARGE\_DATA} &
Does not provide strong evidence for large-fetch behavior; the observed query
outputs are not directly tied to a data-fetch bottleneck. \\
\bottomrule
\end{tabularx}
\end{center}

\end{evidencebox}

\begin{evidencebox}{Judger synthesis over prior analyses}

The judger explicitly refers to the previous agents' investigations rather than
giving an isolated label prediction:

\begin{quote}
\small\itshape
``The previous agents have done some investigations. First, agent1 looked at
\texttt{pg\_stat\_statements} and found a single large insert operation with
48,000 rows. That's a big insert, which could be causing performance issues.
They mentioned that this might be the root cause, pointing to
`\texttt{INSERT\_LARGE\_DATA}'.''
\end{quote}

It then uses other agents' findings as negative evidence against competing
diagnoses:

\begin{quote}
\small\itshape
``Then, agent2 checked for lock contention by querying \texttt{pg\_locks} and
\texttt{pg\_stat\_statements}. They found no active locks, so they ruled out
`\texttt{LOCK\_CONTENTION}'. Agent3 looked at \texttt{pg\_stat\_all\_tables}
and \texttt{pg\_stat\_progress\_vacuum}. They found that the vacuum activity
was minimal, so they didn't see evidence of `\texttt{VACUUM}' being a problem.''
\end{quote}

The judger also notes corroborating support for the large-insert diagnosis:

\begin{quote}
\small\itshape
``Agent5 analyzed the query results and concluded that the large insert
operation was the most likely cause, supporting
`\texttt{INSERT\_LARGE\_DATA}'.''
\end{quote}

Finally, it combines the positive and negative evidence into the final
diagnosis:

\begin{quote}
\small\itshape
``So, putting this all together, the strongest evidence points to
`\texttt{INSERT\_LARGE\_DATA}' as the root cause. The other agents didn't find
significant issues with locks, vacuum, or redundant indexes.''
\end{quote}

Thus the judger outputs:
\[
\texttt{<answer>["INSERT\_LARGE\_DATA"]</answer>}.
\]

\end{evidencebox}

\textbf{Takeaway.}
This example shows that \proj can perform evidence-grounded diagnosis rather
than simply aggregating candidate labels. The judger identifies the strongest
positive signal, a large insert operation involving $48{,}000$ rows, while using
other agents' analyses to rule out competing explanations such as lock
contention, vacuum, and redundant indexes. This leads to a reasonable diagnosis
that matches the gold root cause.

\end{casebox}

\end{document}